# Low-cost commercial LEGO© platform for mobile robotics

Erik Cuevas, Daniel Zaldivar and Marco Pérez-Cisneros

Departamento de Ciencias Computacionales
Universidad de Guadalajara, CUCEI
Av. Revolución 1500, Guadalajara, Jal, México
{erik.cuevas, daniel.zaldivar, marco.perez}@cucei.udg.mx

**Abstract**

This paper shows the potential of a Lego© based low-cost commercial robotic platform for learning and testing prototypes in higher education and research. The overall setup aims to explain mobile robotic issues strongly related to several fields such as Mechatronics, Robotics, and Automatic Control theory. The capabilities and limitations of LEGO© robots are studied within two projects. The first one involves a robotic vehicle which is able to follow several predefined paths. The second project concerns to the classical problem of position control. Algorithms and additional tools have been fully designed, applied and documented with results shown throughout the paper. The platform is found to be suitable for teaching and researching on key issues related to the aforementioned fields.

1. **Introduction**

In the past decades, revolutionary technologies have allowed the development of innovations which have deeply transformed our lives. Mechatronics, a synergistic procedure of mechanical, electronics, software, and control engineering, is breeding a significant contribution to the design and conception of innovative and versatile products. Robotics is becoming more important both for industry and for the design of everyday life. Teaching of Mechatronics and Robotics while focusing on design, optimization, experimentation and development of real-time systems, can be facilitated by the availability of flexible and yet inexpensive rapid prototyping systems.

The well-known LEGO-Lab from Aarhus University [1] and other similar efforts made by St. Brown University [2] and Tufts University [3], have employed LEGO© elements for prototyping and implementation of several ideas related to control, robotics, and computer science disciplines. In addition, student groups from the Department of Automation at the Technical University of Denmark, have created and written code for experimental rigs based the same platform. Such material has been successfully applied to the control systems courses [4]. Handy examples may be taken from Martin [5] which outlines the Mindstorms capabilities while Butler et al. [6] and Lau et al. [7] emphasize the versatility and originality of the LEGO© robots. Lund and Pagliarini created a contest for children known as the Robocup Jr., which aims to build soccer robots using the LEGO© Mindstorm plartform [8]. Recently, Fiorini reported the use of the LEGO components for similar purposes at the University of Verona [9].

At first glance, the LEGO© components offer interesting possibilities. The Mindstorms kits, aimed at advanced users or the Robolab kits, introduced by the Educational Division, include relatively low-cost hardware and software suitable for implementing complex designs for Robotics, Mechatronics and Control. Students can become quickly familiar with the set while taking further steps by adventuring to the designing and testing their own add-on components. They might also interface the kit to other processing and sensing units such as microcontrollers, sensors, etc.

In order to improve the overall learning experience as well as to promote innovative thinking of newborn engineers, new teaching techniques have been based on project designing. Such projects may be either predefined in their scope or free as they focus on creative applications of theoretical class material. In the former case, teams must find reliable and effective solutions employing their creativity and engineering knowledge while learning to overcome practical problems and limitations. In the free project case, additional creativity and innovation are required to define the problem at first hand and work the complete solution on the other hand. For both cases, teams must come up with effective solutions employing both theory and experimentation by means of creative proposals and bounded by feasibility considerations. In addition, such projects inevitably expand student knowledge in areas beyond their immediate interests.

However, the first question to be answered is whether the use of such a set, not initially meant for teaching engineering, may deliver over the teaching requirements related to several fields such as mechanical design, actuators and transmissions, sensor interfacing and/or development, software development, communications design, servo control, etc. Several publications have argued before on the good effect of using LEGO(c) systems in educational context [22]-[24]. In particular, such systems are commonly cheap, robust, reconfigurable, re-programmable and induce enthusiasm and innovation to students [25].

To this end, this work has approached the problem from an engineering point of view. First, key elements on the platform architecture have been modelled to characterize all parameters involved in their operation. In a second step, two relatively complex projects on mobile robotics have been deployed to demonstrate the useful contribution of the platform as partner for teaching. This stage also discusses on the kind of issues to be tackled once the platform is being used.

The first experiment involves a robotic vehicle following a predefined path while the second one concerns to control position problem. The algorithms and all the additional tools are presented together with the simulated and experimental results. The overall advantages and limitations are also discussed. It is suggested that the platform is capable of delivering for training while holding low costs and a friendly learning environment. The paper therefore shows the potential of a Lego© based low-cost commercial robotic platform for learning and testing of experimental prototypes.

The paper is organized as follows. Section 2 presents the overall plant model for the core system for both experiments. Section 3 shows the selected control technique for tracking. Section 4 introduces a regulation technique for position control and its implementation in the ROBOTC compiler. Finally section 5 shows the corresponding conclusions and contributions with respect to the overall aim of the paper.

## 2. Modelling one LEGO© architecture

The mobile unit used throughout this paper is presented by Figure 1(a). The robot is based on the Lego© NXT system. All exercises in this work refer to the differential mobile robot architecture which is characterized on Figure 1(b) following a state-space notation as follows:

$$\begin{bmatrix} \dot{x} \\ \dot{y} \\ \dot{\theta} \end{bmatrix} = \begin{bmatrix} -(c\sin\theta)/2 \\ (c\cos\theta)/2 \\ -c/b \end{bmatrix} \omega_l + \begin{bmatrix} -(c\sin\theta)/2 \\ (c\cos\theta)/2 \\ c/b \end{bmatrix} \omega_r \qquad (1)$$

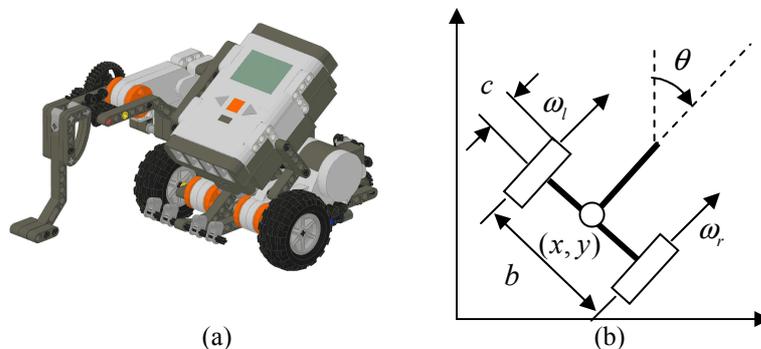

(a)          (b)
**Figure 1.** (a) LEGO(c) platform and (b) Differential kinematics model.

The coordinates of the back shaft´s centre are defined by $(x, y)$, with $\omega_l$ y $\omega_r$ being the left and right wheel velocities respectively, $\theta$ representing the robot´s orientation and $b$ being the length of the back shaft. Following this configuration, *c* represents the wheel's radio which is key parameter within the model. In order to keep the model as simple as possible, a common way is to use a synchronic model which may be represented as follows:

$$\begin{bmatrix} \dot{x} \\ \dot{y} \\ \dot{\theta} \end{bmatrix} = \begin{bmatrix} -\sin\theta & 0 \\ \cos\theta & 0 \\ 0 & 1 \end{bmatrix} \begin{bmatrix} v \\ \omega \end{bmatrix} \qquad (2)$$

By using:

$$v = \frac{(\omega_l + \omega_r)c}{2} \qquad \omega = \frac{(\omega_r - \omega_l)c}{b} \qquad (3)$$

## 3. Trajectory following control using the platform.

### 3.1 Tracker dynamics.

The overall system aims to autonomously guide the robot's movement to track a given trajectory. The control of such problem has been widely studied through the literature, for instance see [20]. The problem can be understood as a simple control law to adjust the robot's position and orientation $\rho = (x, y, \theta)$ in order to minimize the difference to a given target trajectory (see Figure 2) which may be defined by $\rho_{ref} = (x_{ref}, y_{ref}, \theta_{ref})$. Considering a reference model for the trajectory, it may be represented by:

$$\dot{x}_{ref} = -v_{ref}\sin\theta_{ref} \qquad \dot{y}_{ref} = v_{ref}\cos\theta_{ref} \qquad \dot{\theta}_{ref} = \omega_{ref} \qquad (4)$$

So the problem is reduced to consider a control law while assuring the following condition:

$$\lim_{t \to \infty}(\rho(t) - \rho_{ref}(t)) = 0 \qquad (5)$$

Considering Equation 5 and making one required coordinate change, the errors related to the tracking error in the trajectory may be defined as follows:

$$\begin{bmatrix} e_1 \\ e_2 \\ e_3 \end{bmatrix} = \begin{bmatrix} -\sin\theta & \cos\theta & 0 \\ -\cos\theta & -\sin\theta & 0 \\ 0 & 0 & 1 \end{bmatrix} \begin{bmatrix} x_{ref} - x \\ y_{ref} - y \\ \theta_{ref} - \theta \end{bmatrix} \qquad (6)$$

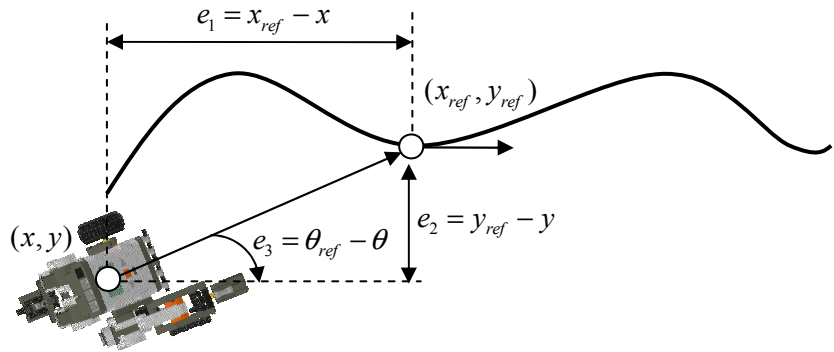

**Figure 2.** Modelling tracking errors and target trajectory.

Therefore the error dynamics may be characterized by:

$$\begin{bmatrix} \dot{e}_1 \\ \dot{e}_2 \\ \dot{e}_3 \end{bmatrix} = \begin{bmatrix} 0 & \omega & 0 \\ -\omega & 0 & 0 \\ 0 & 0 & 1 \end{bmatrix} \begin{bmatrix} e_1 \\ e_2 \\ e_3 \end{bmatrix} + \begin{bmatrix} 0 \\ \sin(e_3) \\ 0 \end{bmatrix} v_{ref} + \begin{bmatrix} 1 & 0 \\ 0 & 0 \\ 0 & 1 \end{bmatrix} \begin{bmatrix} u_1 \\ u_2 \end{bmatrix} \quad (7)$$

With

$$u_1 = -v + v_{ref} \cos(e_3) \qquad u_2 = \omega_{ref} - \omega \quad (8)$$

Further analyzing (7), it yields

$$\begin{bmatrix} \dot{e}_1 \\ \dot{e}_2 \\ \dot{e}_3 \end{bmatrix} = \begin{bmatrix} 0 & \omega_{ref} & 0 \\ -\omega_{ref} & 0 & v_{ref} \\ 0 & 0 & 1 \end{bmatrix} \begin{bmatrix} e_1 \\ e_2 \\ e_3 \end{bmatrix} + \begin{bmatrix} 1 & 0 \\ 0 & 0 \\ 0 & 1 \end{bmatrix} \begin{bmatrix} u_1 \\ u_2 \end{bmatrix} \quad (9)$$

In [20], a simple control law is proposed as

$$u_1 = -k_1 e_1 \qquad u_2 = -k_2 \operatorname{sgn}(v_{ref}) e_2 - k_3 e_3 \quad (10)$$

Taking (10) into (9), it yields a composite expression of the error dynamics within the trajectory following problems leaving

$$\begin{bmatrix} \dot{e}_1 \\ \dot{e}_2 \\ \dot{e}_3 \end{bmatrix} = \begin{bmatrix} -k_1 & \omega_{ref} & 0 \\ \omega_{ref} & 0 & v_{ref} \\ 0 & -k_2 \operatorname{sgn}(v_{ref}) & -k_3 \end{bmatrix} \begin{bmatrix} e_1 \\ e_2 \\ e_3 \end{bmatrix} \quad (11)$$

The characteristic roots are obtained from the polynomial $\det(s\mathbf{I} - \mathbf{A})$ built from (11), yielding

$$(s + k_1)[s(s + k_1) + v_{ref} k_2 \operatorname{sgn}(v_{ref}) + \omega_{ref}^2] = 0 \quad (12)$$

Thus, (12) can be written as

$$(s + 2\xi b)[s^2 + 2\xi bs + b^2] = 0 \quad (13)$$

In this case, $\xi$ represents the damping factor and $b$ the natural frequency for the dynamics represented by (10). In order to control the dynamics of the tracking, constants $k_1$, $k_2$ y $k_3$ must be determined following

$$\begin{aligned} k_1 &= 2\xi b \\ k_2 &= \frac{b^2 - \omega_{ref}^2}{|v_{ref}|} \\ k_3 &= 2\xi b \end{aligned} \quad (14)$$

An interested reader may find a deeper treatment of these results in [20].

## 4. Position Control

### 4.1 Positioning System Dynamics

The overall aim in the position control problem is to find a control law capable of translating the LEGO© robot from an initial position to the target. Considering the example shown in Figure 4, the mobile robot is located at a position **I**, $P_I = (x_I, y_I, \theta_I)$ and it may be translated to position **E** which in turn may be defined as $P_E = (x_E, y_E, \theta_E)$.

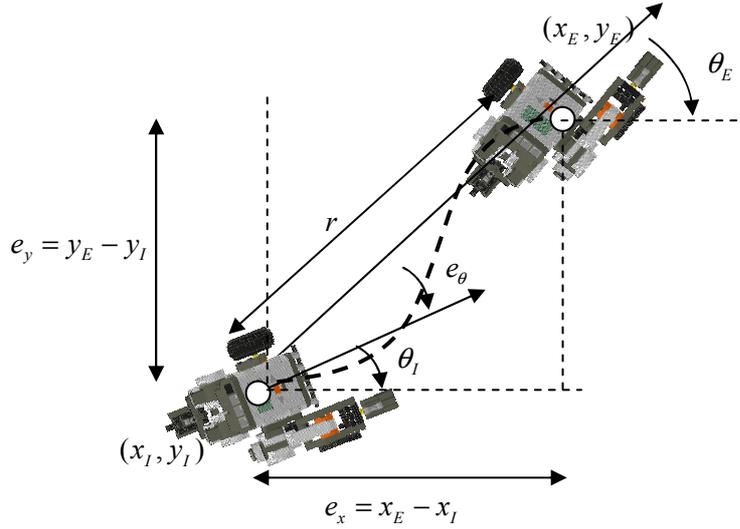

**Figure 4.** Position control problem for a mobile robot.

Therefore, it is required to calculate matrix $K$ as follows

$$K = \begin{bmatrix} k_{11} & k_{12} & k_{13} \\ k_{21} & k_{22} & k_{23} \end{bmatrix} \text{ with } k_{ij} = k(t,e) \tag{15}$$

Through computation of signals $v(t)$ and $\omega(t)$ and according to (2), they may control the overall behaviour of the mobile robot following

$$\begin{bmatrix} v(t) \\ \omega(t) \end{bmatrix} = K \cdot e \tag{16}$$

So the error $e$ tends to zero and therefore $\lim_{t \to \infty}(e(t)) = 0$. Thus, defining all system's variables according to Figure 4, they yield

$$\begin{aligned} r &= \sqrt{e_1^2 + e_2^2} = \sqrt{(x_E - x_I)^2 + (y_E - y_I)^2} \\ e_\theta &= -\theta_I + \text{atan2}(e_2, e_1) \\ \theta_E &= -\theta_I - e_\theta \end{aligned} \tag{17}$$

Linking all these definitions to (2), the dynamical system may also be defined as

$$\begin{bmatrix} \dot{r} \\ \dot{e}_\theta \\ \dot{\theta}_E \end{bmatrix} = \begin{bmatrix} -\cos(e_\theta) & 0 \\ \dfrac{\sin(e_\theta)}{r} & -1 \\ -\dfrac{\sin(e_\theta)}{r} & 0 \end{bmatrix} \begin{bmatrix} v \\ \omega \end{bmatrix} \tag{18}$$

Considering (18), the overall problem is focus on defining signals $v(t)$ y $\omega(t)$ which may take the robot from its initial position $P_I = (x_I, y_I, \theta_I)$ to its target in $P_E = (x_E, y_E, \theta_E)$. By using the control law previously defined from [21], it yields

$$\begin{aligned} v &= k_r r \\ \omega &= k_{e_\theta} e_\theta + k_{\theta_E} \theta_E \end{aligned} \tag{19}$$

Substituting in (18), the overall dynamics for the mobile robot is now defined as follows

$$\begin{bmatrix} \dot{r} \\ \dot{e}_\theta \\ \dot{\theta}_E \end{bmatrix} = \begin{bmatrix} -k_r \cdot r \cdot \cos(e_\theta) \\ k_r \cdot \sin(e_\theta) - k_{e_\theta} \cdot e_\theta - k_{\theta_E} \cdot \theta_E \\ -k_r \cdot \sin(e_\theta) \end{bmatrix} \tag{20}$$

Applying linearization over expression (20) by means of typical trigonometric approximations such as $\cos(x) = 1$ and $\sin(x) = x$, it yields

$$\begin{bmatrix} \dot{r} \\ \dot{e}_\theta \\ \dot{\theta}_E \end{bmatrix} = \begin{bmatrix} -k_r & 0 & 0 \\ 0 & -(k_{e_\theta} - k_r) & -k_{\theta_E} \\ 0 & -k_r & 0 \end{bmatrix} \begin{bmatrix} r \\ e_\theta \\ \theta_E \end{bmatrix} \tag{21}$$

The characteristic roots may be also obtained from the polynomial $\det(s\mathbf{I} - \mathbf{A})$ transforming expression (21) into

$$(s + k_r)(s^2 + s(k_{e_\theta} - k_r) - k_r k_{\theta_E}) = 0 \tag{22}$$

From expression (22), it is evident that gains $k_r$, $k_{e_\theta}$ y $k_{\theta_E}$ must be limited to the subspaces in expression (23) in order to keep the overall system stability.

$$\begin{aligned} k_r &> 0 \\ k_{\theta_E} &< 0 \\ k_{e_\theta} - k_r &> 0 \end{aligned} \tag{23}$$

A deeper study on the subject may be found in [21].

### 4.2 Implementation in ROBOTC language

Different languages and compilers for the LEGO© platform are currently available. They provide multiple options to overcome the commonly-found constraints while developing innovative robotic exercises under a friendly and easy to execute code.

In the particular case of control engineering, real-time systems require floating point operations, extended sets of mathematical functions and clear debugging tools. The ROBOTC© compiler is therefore chosen as a convenient platform which delivers solutions to such requirements. However, it also shows some important compiling problems to be solved through a neat and comprehensive use of debugging tools and useful programming concepts.

For instance, implementing the system on equation 17, requires multiple calls to the atan2( ) trigonometric function. ROBOTC© does not support this type of function and it has to be implemented within the algorithm´s core. Such atan2 function can be found in the pseudocode section below.

Another important problem is related to the fact that the ROBOTC© compiler cannot implement functions with multiple operators. So, for such a case, operators should be distributed along several lines using an extra set of temporal variables.

An important problem arises from the fact that magnitudes of the calculated values resulting from Equation 19, cannot be supported by the NXT architecture, in particular power in each of the motors. The maximum value within the NXT system is ±100 units. However, the control signal –particularly at the very beginning, exceed this value. If unaccounted, this situation might generate unexpected results and overall system´s instability.

The problem has been solved in this paper by using a factor to multiply each control variable $v$ and $\omega$ from Equation 19. Such factor then grows exponentially from 0 to 1. Therefore, it must guarantee that all computed values do not exceed the limits supported by the hardware. Such factor is defined as:

$$1 - e^{-\alpha t} \qquad (24)$$

The parameter $\alpha$ can be adjusted to generate appropriate responses. Considering that the robot goes from the point (0,0) to the point (1,1), the initial angle is $\pi/2$ and the final angle is $\pi$. Figures 5 and 6 show the differences between the control without exponential modification and the enhanced control implemented within the NXT system.

```
% initialize global variables
%Position of the robot:
    x_I IS 0
    y_I IS 0
%Final position:
    x_E IS (desired x position in meters)
    y_E IS (desired y position in meters)
%Angle:
theta IS (angle of the robot in radians)
%Final Angle:
    bo IS (desired angle in radians)
%Robot Constants:
    wheel_ratio IS 0.0275
    axis IS 0.135
%Time:
    t IS 0
    time_step IS 0.03
%Odometry
    left_encoder IS 0
    right_encoder IS 0

%control loop:
WHILE position of the robot IS NOT final position
    CALL_FUNCT xyt()
    xd IS xo – x  "distance in x axis
    yd IS yo – y  "distance in y axis
%distance vector
    r IS  √((x_E − x_I)² + (y_E − y_I)²)
```

```
%alpha→ e_θ
    alpha IS -theta + CALL FUNCT atan2(yd,xd)
%beta→ θ_E
    beta IS -theta - alpha + bo
%r_gain→ k_r
% small at the beginning and big at the end
    r_gain IS  0.4(1−e^{−0.1t})
    alpha_gain IS  1.4(1−e^{−0.1t})
    beta_gain IS  1.5(1−e^{−0.3t})
%linear velocity
    speed IS r_gain * r
%angular velocity
ang_speed IS (alpha_gain * alpha) + (beta_gain * beta)

%angular velocity of each wheel
wheel_ratio IS wR
left_ang_speed IS (speed - (axis/2) * ang_speed) / wR
right_ang_speed IS (speed + (axis/2) * ang_speed) / wR

%angular speeds of each wheel in degrees per second.
    wl IS (left_ang_speed * 57.2957) * 0.1010 + 0.4372
    wr IS (right_ang_speed * 57.2957) * 0.1010 + 0.4372

%power on the motors of the robot
    motor[left_motor] IS wl
    motor[right_motor] IS wr
```

```
% waits 30 milliseconds to get enough pulses of
% the "encoders
    WAIT(30msecs)
%time increment
    t IS t + time_step
%end of control loop
END WHILE
```

**Algorithm 1.** Postion controller as implemented on ROBOTC©.

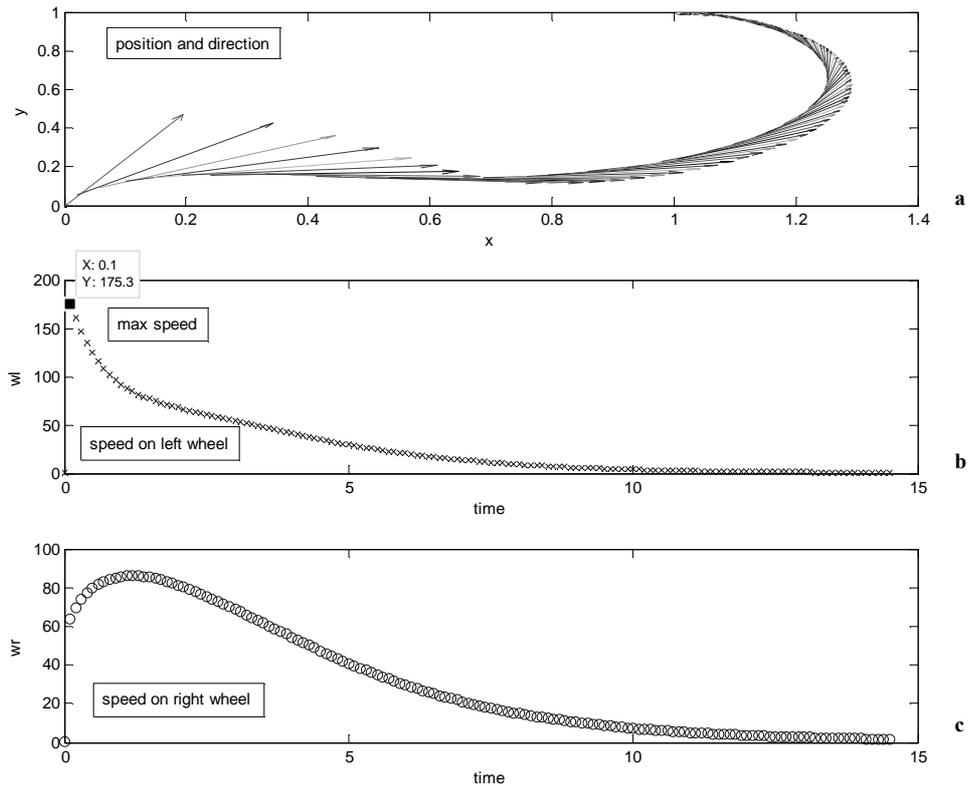

**Figure 5.**- Applied control with no exponential factor. (a) Velocity vector while the robot moves from (0,0) to (1,1), (b) Left wheel´s angular velocity, and (c) angular velocity on the right tire. Gain values are defined as $k_r = 0.4, k_{e_\theta} = 2$ y $k_{\theta_E} = -1$.

Figure 5 shows the applied control with no exponential factor. Figure 6 presents the controller as implemented for the NXT system but including the exponential value. By adding such factor, the control value applied to the motor is reduced while the control response becomes slower. This is a required effect given the risk of getting instability once the system is operating the robot in real-time.

### 4.2.1 Writing the algorithm using ROBOTC©

Considering expressions 17 and 19, it is possible to draw the overall control algorithm, including the constraint set to assure fully compliance of the conditions shown by Equation 23. Algorithm 1, shows the overall procedure by means of the RobotC© language.

```
FUNCT atan2(yd,xd)
   IF yd IS 0 THEN

      IF xd IS GREATER_THAN 0 THEN
        RETURN 0
      END IF

   IF xd IS 0 THEN
        RETURN 0
      END IF

      IF xd IS LESS THAN 0 THEN
        RETURN pi
      END IF
   ELSE
      IF xd IS GREATER_THAN 0 THEN
         angle IS ATAN(yd/xd)
         RETURN angle
      END IF
```

```
%SIGN_OF returns -1 or 1 depending on the sign
%of the argument
         IF xd IS 0 THEN
            angle IS (pi/2) * SIGN_OF(yd)
            RETURN angle
         END IF

         IF xd IS LESS_THAN 0 THEN
   angle IS ATAN(yd/xd) + [pi * SIGN_OF(yd)]
            RETURN angle
         END IF

     END IF
END FUNCT
```

**Algorithm 2.** Source code for the function `atan2(yd,xd)`.

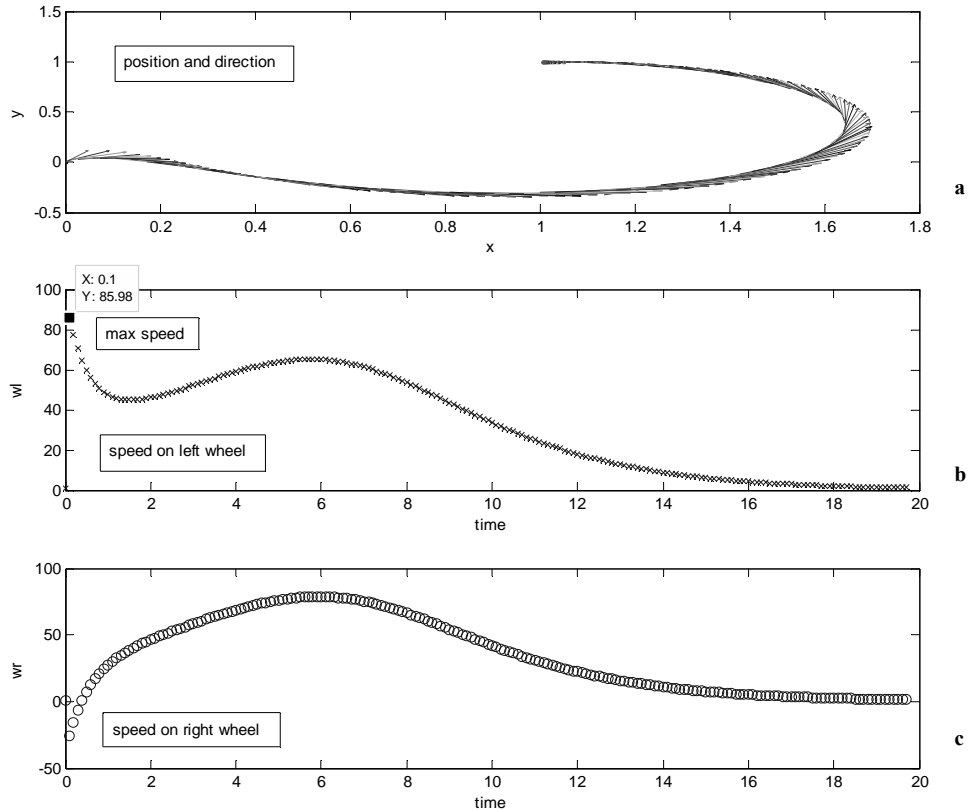

**Figure 6.** Applied control using the NXT set including the exponential factor. (a) Velocity vector while the robot moves from (0,0) to (1,1), (b) Left wheel angular velocity, and (c) angular velocity on the right tire. Gain values are defined as $k_r = 0.4(1-e^{-0.1t}), k_{e_\theta} = 2(1-e^{-0.1t})$ y $k_{\theta_E} = -1(1-e^{-0.3t})$.

Within the algorithm 1, there are two functions: `FUNCT xyt()` and `FUNCT atan2(yd,xd)`. The former expression is an easy implementation which allows to read and convert encoder values into the corresponding (*x,y*) coordinates. On the other hand, the function `FUNCT atan2(yd,xd)`, allows to implement the atan( ) function which is not included in the function list within the compiler. The algorithm table 2 describes such function.

## 5. Conclusions

This paper shows the potential of a Lego© NXT based low-cost commercial robotic platform for learning and testing prototypes in higher education and research. The study implements two control algorithms for navigating mobile robots based on the NTX platform. The discussion includes an introduction to NXT features and to the ROBOTC© programming learning which in turn is used to implement trajectory control and position control on the mobile robot.

The ROBOTC© has been chosen in particular thanks to its ability to deal with special non-supported functions in order to solve challenging problems during the designing stage. Therefore it is possible to generate friendly compact code which delivers a simple and easy to understand implementation of each controller over the NXT platform.

Saturation problems exhibited by servo-motors can be solved by subtle adaptations to the original algorithm. In this paper, an exponential gain was conceived to reduce actuator's saturation. The effect of applying such gain, after a proper parameter calibration, may be irrelevant for the final implementation of both controllers.